\documentclass{article}

\usepackage[final, nonatbib]{nips_2016}

\usepackage[utf8]{inputenc} 
\usepackage[T1]{fontenc}    
\usepackage{hyperref}       
\usepackage{url}            
\usepackage{booktabs}       
\usepackage{amsfonts}       
\usepackage{nicefrac}       
\usepackage{microtype}      
\usepackage{bm}
\usepackage{color}
\usepackage{csquotes}
\usepackage{graphicx}
\usepackage{amssymb}
\usepackage{amsmath}
\usepackage{latexsym}
\usepackage{algpseudocode}
\usepackage{algorithm}
\usepackage[inline]{enumitem}
\newcommand{\quotes}[1]{``#1''}

\usepackage{tikz}
\usepackage{pgfplots}
\pgfplotsset{compat=1.9}
\usepackage{verbatim}
\usepackage{textcomp}
\usetikzlibrary{shapes,arrows,positioning}
\usepackage[normalem]{ulem}

\title{Neural Networks Regularization Through Class-wise Invariant Representation Learning}

\author{
  Soufiane Belharbi\thanks{https://sbelharbi.github.io} \\
  Normandie Univ, UNIROUEN, UNIHAVRE,\\
  INSA Rouen, LITIS\\
  76000 Rouen, France \\
  \texttt{soufiane.belharbi@insa-rouen.fr} \\
  \And
  Clément Chatelain \\
  Normandie Univ, UNIROUEN, UNIHAVRE,\\
  INSA Rouen, LITIS\\
  76000 Rouen, France \\
  \texttt{clement.chatelain@insa-rouen.fr} \\
  \AND
  Romain Hérault\\
  Normandie Univ, UNIROUEN, UNIHAVRE,\\
  INSA Rouen, LITIS\\
  76000 Rouen, France \\
  \texttt{romain.herault@insa-rouen.fr} \\
  \And
  Sébastien Adam\\
  Normandie Univ, UNIROUEN, UNIHAVRE,\\
  INSA Rouen, LITIS\\
  76000 Rouen, France \\
  \texttt{sebastien.adam@univ-rouen.fr} \\
}

\begin{document}

\maketitle

\begin{abstract}
Training deep neural networks is known to require a large number of training samples. However, in many applications only few training samples are available. In this work, we tackle the issue of training neural networks for classification task when few training samples are available. We attempt to solve this issue by proposing a new regularization term that constrains the hidden layers of a network to learn \emph{class-wise invariant representations}. In our regularization framework, learning invariant representations is generalized to the class membership where samples with the same class should have the same representation. Numerical experiments over MNIST and its variants showed that our proposal helps improving the generalization of neural network particularly when trained with few samples. We provide the source code of our framework \footnote{\url{https://github.com/sbelharbi/learning-class-invariant-features}}.
\end{abstract}

\section{Introduction}
\label{intro}
For a long time, it has been understood in the field of deep learning that building a model by stacking multiple levels of non-linearity is an efficient way to achieve good performance on complicated artificial intelligence tasks  such as vision \cite{ krizhevsky12, simonyanZ14a, szegedyLJSRAEVR14, heZRS16} or natural language processing \cite{CollobertWeston2008, WestonRC2008, Kim14, Graves13}. The rationale behind this statement is the hierarchical learned representations throughout the depth of the network which circumvent the need of extracting handcrafted features.

For many years, the non-convex optimization problem of learning a neural network has prevented going beyond one or two hidden layers. In the last decade, deep learning has seen a breakthrough with efficient training strategies of deeper architectures\cite{hinton06NC, RanzatoPCL2006, bengio06NIPS}, and a race toward deeper models has began\cite{krizhevsky12, simonyanZ14a, szegedyLJSRAEVR14, heZRS16}. This urge to deeper architectures was due to (i) large progress in optimization, (ii) the powerful computation resources brought by GPUs\footnote{Graphical Processing Units.} and (iii) the availability of huge datasets such as ImageNet \cite{imagenet09} for computer vision problems. However, in real applications, few training samples are usually available which makes the training of deep architectures difficult. Therefore, it becomes necessary to provide new learning schemes for deep networks to perform better using few training samples. 

A common strategy to circumvent the lack of annotated data is to exploit extra informations related to the data, the model or the application domain, in order to guide the learning process. This is typically carried out through regularization which can rely for instance on data augmentation, $L_2$ regularization \cite{tikhonov77}, dropout\cite{srivastava14a}, unsupervised training \cite{hinton06NC, RanzatoPCL2006, bengio06NIPS, rifaiVMGB11, salah2011, bel16}, shared parameters\cite{leCun1989, riesenhuber99, fukushimaM82},  etc. 

Our research direction in this work is to provide a new regularization framework to guide the training process in a supervised classification context. The framework relies on the exploitation of prior knowledge which has already been used in the literature to train and improve models performance when few training samples are available  \cite{mitchell1997, scholkopf2001, heckerman1995, niyogi98, krupka2007, yu2007, wu2004, YuSJ10Neurocomp}.

Indeed, prior knowledge can offer the advantage of more consistency, better generalization and fast convergence using less training data by guiding the learning process \cite{mitchell1997}. By using prior knowledge about the target function, the learner has a better chance to generalize from sparse data \cite{mitchell1997, abuMostafa90, abuMostafab92,abuMostafa93}. For instance, in object localization such as part of the face, knowing that the eyes are located above the nose and the mouth can be helpful. One can exploit this prior structure about the data representation: to constrain the model architecture, to guide the learning process, or to post-process the model's decision.

In classification task, although it is difficult to define what makes a representation  good, two properties are inherent to the task: \emph{Discrimination} i.e. representations must allow to separate samples of distinct classes. \emph{Invariance} i.e. representations must allow to obtain robust decision despite some variations of input samples. Formally, given two samples $x_1$ and $x_2$, a representation function ${\Gamma(\cdot)}$ and a decision function $\Psi(\cdot)$;  when ${x_1 \approx x_2}$ , we seek invariant representations that provide ${\Gamma(x_1) \approx \Gamma(x_2)}$, leading to smooth decision ${\Psi(\Gamma(x_1)) \approx\Psi(\Gamma(x_2))}$. In this work, we are interested in the invariance aspect of the representations. This definition can be extended to more elaborated transformations such as rotation, scaling, translation, etc. However, in real life there are many other transformations which are difficult to formalize or even enumerate. Therefore, we extend in this work the definition of the invariant representations to the class membership, where samples within the same class should have the same representation. At a representation level, this should generate homogeneous and tighter clusters per class.

\begin{figure*}
	\centering
\resizebox{\textwidth}{!}{\input{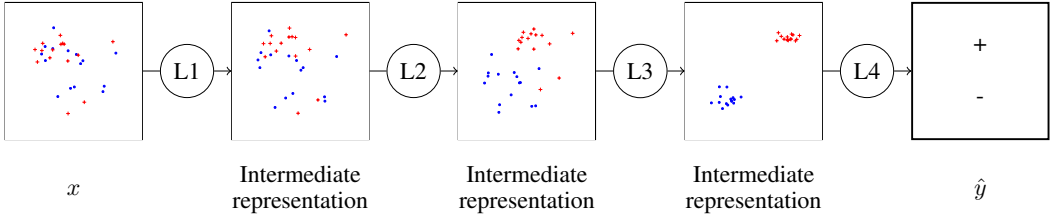}}
\caption{Input/Hidden representations of samples from an artificial dataset along  4 layers of a MLP. Each representation is projected into a 2D space. }
\label{fig:datarep}
\end{figure*}

In the training of neural networks, while the output layer is guided by the provided target, the hidden layers are left to the effect of the propagated error from the output layer without a specific target.
Nevertheless, once the network trained, examples may form (many) modes on hidden representations, i.e. outputs of hidden layers, conditionally to their classes. Most notably, on the penultimate representation before the decision stage, examples should agglomerate in distinct clusters according to their label as seen on Figure \ref{fig:datarep}.
From the aforementioned prior perspective about the hidden representations, we aim in this work to provide a learning scheme that promotes the hidden layers to build representations which are class-invariant and thus agglomerate in restricted number of modes. By doing so, we constrain the network to build invariant intermediate representations per class with respect to the variations in the input samples \emph{without explicitly specifying these variations nor the transformations that caused them}. 

We express this class-invariance prior as an explicit criterion combined with the classification training criterion. It is formulated as a dissimilarity between the representations of each pair of samples within the same class. The average dissimilarity over all the pairs of all the classes is considered to be minimized. To the best of our knowledge, none has used this class membership to build invariant representations. Our motivation in using this prior knowledge, as a form of regularization, is to be able to train deep neural networks and obtain better generalization error using less training data. We have conducted different experiments over MNIST benchmarck using two models (multilayer perceptrons and convolutional networks) for different classification tasks. We have obtained results that show important improvements of the model's generalization error particularly when trained with few samples.

The rest of the paper is organized as follows: Section \ref{related} presents related works for invariance learning  in neural networks. We present our learning framework in Section \ref{method} followed by a discussion of the obtained results in Section \ref{exps}.

\section{Related Work}
\label{related}
Learning general invariance, particularly in deep architectures, is an attractive subject where different approaches have been proposed. The rational behind this framework is to ensure the invariance of the learned model toward the variations of the input data. In this section, we describe three kinds of approaches of learning invariance within neural networks. Some of these methods were not necessarily designed to learn invariance however we present them from the invariance perspective. For this description, $f$ is the target function to be learned.

\begin{description}

\item [\emph{Invariance through data transformations}:]
\hfill \\
It is well known that generalization performance can be improved by using larger quantity of training samples. Enlarging the number of samples can be achieved by generating new samples through the application of small random transformations such as rotation, scaling, random noise, etc \cite{Baird90, Ciresan2010, Simard2003} to the original examples. Incorporating such transformed data within the learning process has shown to be helpful in generalization \cite{niyogi98}. \cite{abuMostafa90} proposes the use of prior information about the behavior of $f$ over perturbed examples using different transformations where $f$ is constrained to be invariant over all the samples generated using these transformations. While data transformations successfully incorporate certain invariance into the learned model, they remain limited to some predefined and well known transformations. Indeed, there are many other transformations which are either unknown or difficult to formalize.

\item [\emph{Invariance through model architectures}:]
\hfill \\
In some neural network models, the architecture implicitly builds a certain type of invariance. For instance, in convolutional networks \cite{leCun1989, riesenhuber99, fukushimaM82}, combining layers of feature extractors using weight sharing with local pooling of the feature maps introduces some degree of translation invariance \cite{ranzato-cvpr-07, lee2009}. These models are currently state of the art strategies for achieving invariance in computer vision tasks. However, it is unclear how to explicitly incorporate in these models more complicated invariances such as large angle rotation and complex illumination. Moreover, convolutional and max-pooling techniques are somewhat specialized to visual and audio processing, while deep architectures are generally task independent.

\item [\emph{Invariance through analytical constraints}:]
\hfill \\
Analytical invariance consists in adding an explicit penalty term to the training objective function in order to reduce the variations of $f$ or its sub-parts when the input varies. This penalty is generally based on the derivatives of a criterion related to $f$ with respect to the input.
For instance, in unsupervised representation learning, \cite{rifaiVMGB11} introduces a penalty for training auto-encoders which encourages the intermediate representation to be robust to small changes of the input around the training samples, referred to as contractive auto-encoders. This penalty is based on the Frobenius norm of the first order derivative of the hidden representation of the auto-encoder with respect to the input. Later, \cite{salah2011} extended the contractive auto-encoders by adding another penalty using the norm of an approximation of the second order derivative of the hidden representation with respect to the input. The added term penalizes curvatures and thus favors smooth manifolds. \cite{shixiang2014} exploit the idea that solving adversarial examples is equivalent to increase the attention of the network to small perturbation for each example. Therefore, they propose a layer-wise penalty which creates flat invariance regions around the input data using the contractive penalty proposed in \cite{rifaiVMGB11}. \cite{simVicLeCDen92, simard1993} penalize the derivatives of $f$ with respect to perturbed inputs using simple distortions in order to ensure local invariance to these transformations.
Learning invariant representations through the penalization of the derivatives of the representation function $\Gamma(\cdot)$ is a strong mathematical tool. However, its main drawback is that the learned invariance is local and is generally robust toward small variations.

\bigskip

Learning invariance through explicit analytical constraints can also be found in metric learning. For instance, \cite{chopraHL05, hadsellCL06} use a contrastive loss which constrains the projection in the output space as follows: input samples annotated as \emph{similar} must have close (adjacent) projections and samples annotated as \emph{dissimilar} must have far projections. In the same way, Siamese networks\cite{bromley1993} proceed in learning similarity by projecting input points annotated as \emph{similar} to be adjacent in the output space. This approach of analytical constraints is our main inspiration in this work, where we provide a penalty that constrains the representation function $\Gamma(\cdot)$ to build similar representation for samples from the same class, i.e. in a supervised way. 
\end{description}
\bigskip

In the following section, we present our proposal with more details.

\section{Proposed Method}
\label{method}
In deep neural networks, high layers tend to learn abstract representations that we have assumed to be closer and closer for the same class along the layers. We would like to promote this behavior. In order to do so, we add a penalty to the training criterion of the network to constrain intermediate representations to be class-invariant. We first describe a model decomposition, the general training framework and then more specific implementation details.

\subsection{Model Decomposition}

Let us consider a parametric mapping function for classification: ${\mathcal{M}(.; \theta): \mathbf{X} \to \mathbf{Y}}$, represented here by a neural network model, where $\mathbf{X}$ is the input space and $\mathbf{Y}$ is the label space.
This neural network is arbitrarily decomposed into two parametric  sub-functions: 
\begin{enumerate*}
  \item ${\Gamma(\cdot;\theta_{\Gamma}): \mathbf{X} \to \mathbf{Z}}$, a representation function parameterized with the set $\theta_{\Gamma}$. This sub-function projects an input sample $x$ into a representation space $\mathbf{Z}$.
  \item ${\Psi(\cdot;\theta_{\Psi}): \mathbf{Z} \to \mathbf{Y}}$, a decision function parameterized with the set $\theta_{\Psi}$. It performs the classification decision over the representation space $\mathbf{Z}$.
\end{enumerate*}

The network decision function can be written as follows:
\begin{equation}
\label{eq:eq00}
\mathcal{M}(x_i;\theta) = \Psi(\Gamma(x_i;\theta_{\Gamma}); \theta_{\Psi})
\end{equation}
\noindent where  ${\theta=\{\theta_{\Gamma},\theta_{\Psi}\}}$.

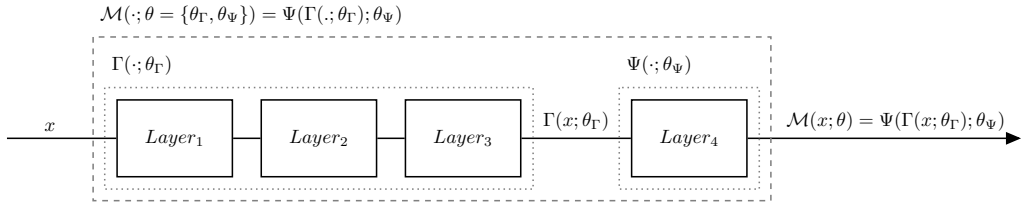
\begin{figure*}[ht]
	\centering
	\resizebox{\textwidth}{!}{

\tikzset{%
  block/.style    = {draw, thick, rectangle, minimum height = 4em,
    minimum width = 6em}
}

\begin{tikzpicture}[auto, thick, >=triangle 45]
\draw
    node at (0,0) (xin) {}
    node [block, right=20mm of xin.east, anchor=west] (l1) {$Layer_1$}
    node [block, right=5mm of l1.east, anchor=west] (l2) {$Layer_2$}
    node [block, right=5mm of l2.east, anchor=west] (l3) {$Layer_3$}
    node [block, right=20mm of l3.east, anchor=west] (l4) {$Layer_4$}
    node [right=50mm of l4.east, anchor=west ] (yout) {}
    ;
    
    \path (l3)--(l4)  node[pos=0.5]{$\Gamma(x;\theta_\Gamma)$};

    \draw[-](xin) -- node {} (l1);
    \draw[-](l1) -- node {} (l2);
    \draw[-](l2) -- node {} (l3);
    \draw[-](l3) -- node {} (l4);
    \draw[->](l4) -- node {} (yout);
  
    \coordinate[above left=3mm of l1.north west] (gammanw);
    \coordinate[below right=3mm of l3.south east] (gammase);
    \draw [color=gray,thick, dotted] (gammanw) rectangle (gammase);
 	\node[above=1mm of gammanw, anchor=south west]  (gammalabel) {$\Gamma(\cdot;\theta_\Gamma)$};
  
    \coordinate[above left=3mm of l4.north west] (psinw);
	\coordinate[below right=3mm of l4.south east] (psise);
	\draw [color=gray,thick, dotted] (psinw) rectangle (psise);
	\node[above=1mm of psinw, anchor=south west]  {$\Psi(\cdot;\theta_\Psi)$};
 
  \coordinate[above left=3mm of gammalabel.north west] (mnw);
  \coordinate[below right=3mm of psise] (mse);
  \draw [color=gray,thick, dashed] (mnw) rectangle (mse);
  \node[above=1mm of mnw, anchor=south west]  {$\mathcal{M}(\cdot;\theta=\{\theta_{\Gamma},\theta_{\Psi}\})= \Psi(\Gamma(.;\theta_{\Gamma});\theta_{\Psi})$};
  
  \path (mnw|-xin)--(xin) node[pos=0.5,above] {$x$};
  \path (mse|-yout)--(yout) node[pos=0.5,above] {$\mathcal{M}(x;\theta)=\Psi(\Gamma(x;\theta_\Gamma);\theta_\Psi)$};

\end{tikzpicture}}
	\caption{Decomposition of the neural network $\mathcal{M}(\cdot)$ into a representation function $\Gamma(\cdot)$ and a decision function $\Psi(\cdot)$.}
	\label{fig:fig0-1}
\end{figure*}

Such a possible  decomposition of a neural network with $K=4$ layers is presented in Fig.\ref{fig:fig0-1}. Here, the decision function $\Psi(\cdot)$ is composed of solely the output layer while the rest of the hidden layers form the representation function $\Gamma(\cdot)$.

\subsection{General Training Framework}

In order to constrain the intermediate representations $\Gamma(\cdot)$ to form clusters over all the samples within the same class we modify the training loss by adding a regularization term. Thus, the training criterion $J$ is composed of the sum of two terms. The first term $\bm{J_{sup}}$ is a standard supervised term which aims at reducing the classification error. The second and proposed regularization term $\bm{J_{H}}$ is a \emph{hint penalty} that aims at constraining the intermediate representations of samples within the same class to be similar. 
By doing so, we constrain $\Gamma(\cdot)$ to lean invariant representations with respect to the class membership of the input sample.

\textbf{Proposed Hint Penalty}

Let ${\mathcal{D} = \{(x_i, y_i)\}}$ be a training set for classification task with $S$ classes and $N$ samples; $(x_i, y_i)$ denotes an input sample and its label. Let $\mathcal{D}_s$ be the sub-set of $\mathcal{D}$ that consists in all the examples of class $s$, i.e. $\mathcal{D}_s=\{(x,y) \in \mathcal{D}\enspace s.t. \enspace y=s\}$.
By definition, $\mathcal{D}=\bigcup\limits_{s=1}^S\mathcal{D}_s$.
For the sake of simplicity,  even if $\mathcal{D}$ and  $\mathcal{D}_s$ contains tuples of (feature,target),  $x$ represents only the feature part in the notation $x \in \mathcal{D}$.

Let $x_i$ be an input sample. We want to reduce the dissimilarity over the space $\mathbf{Z}$ between the projection of $x_i$ and the projection of every sample ${x_j \in \mathcal{D}_s}$ with $j\neq{i}$. 
For this sample $x_i$, our hint penalty can be written as follows:

\begin{equation}
  \label{eq:eq0}
  J_{h}(x_i; \theta_{\Gamma}) = \frac{1}{|\mathcal{D}_s|-1}\sum_{\substack{{x_j \in \mathcal{D}_s}\\{j \neq i}}} \mathcal{C}_h(\Gamma(x_i;\theta_{\Gamma}), \Gamma(x_j; \theta_{\Gamma}))
\end{equation}
\noindent where $\mathcal{C}_{h}(\cdot, \cdot)$ is a loss function that measures how much two projections in $\mathbf{Z}$ are dissimilar and $|\mathcal{D}_s|$ is the number of samples in $\mathcal{D}_s$.

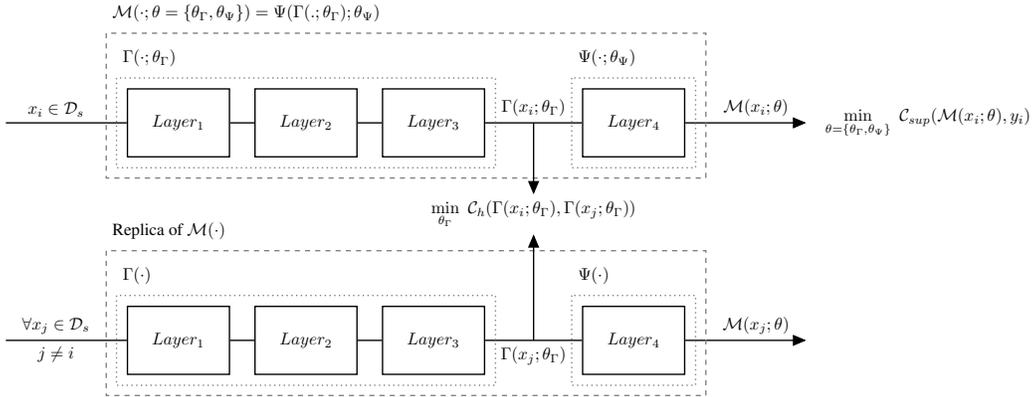
\begin{figure*}[ht!]
\centering
\resizebox{\textwidth}{!}{

\tikzset{%
  block/.style    = {draw, thick, rectangle, minimum height = 4em,
    minimum width = 6em}
}
\begin{tikzpicture}[auto, thick, >=triangle 45]

\draw
node at (0,0) (xin) {}
node [block, right=25mm of xin.east, anchor=west] (l1) {$Layer_1$}
node [block, right=5mm of l1.east, anchor=west] (l2) {$Layer_2$}
node [block, right=5mm of l2.east, anchor=west] (l3) {$Layer_3$}
node [block, right=20mm of l3.east, anchor=west] (l4) {$Layer_4$}
node [right=25mm of l4.east, anchor=west ] (yout) {}
;
\path (l3)--(l4)  coordinate[pos=0.5] (gammai) node[pos=0.5]{$\Gamma(x_i;\theta_{\Gamma})$};

\draw[-](xin) -- node {} (l1);
\draw[-](l1) -- node {} (l2);
\draw[-](l2) -- node {} (l3);
\draw[-](l3) -- node {} (l4);
\draw[->](l4) -- node {} (yout);

\coordinate[above left=3mm of l1.north west] (gammanw);
\coordinate[below right=3mm of l3.south east] (gammase);
\draw [color=gray,thick, dotted] (gammanw) rectangle (gammase);
\node[above=1mm of gammanw, anchor=south west]  (gammalabel) {$\Gamma(\cdot;\theta_{\Gamma})$};

\coordinate[above left=3mm of l4.north west] (psinw);
\coordinate[below right=3mm of l4.south east] (psise);
\draw [color=gray,thick, dotted] (psinw) rectangle (psise);
\node[above=1mm of psinw, anchor=south west]  {$\Psi(\cdot;\theta_{\Psi})$};

\coordinate[above left=3mm of gammalabel.north west] (mnw);
\coordinate[below right=3mm of psise] (msei);
\draw [color=gray,thick, dashed] (mnw) rectangle (msei);
\node[above=1mm of mnw, anchor=south west]  {$\mathcal{M}(\cdot;\theta=\{\theta_{\Gamma},\theta_{\Psi}\})= \Psi(\Gamma(.;\theta_{\Gamma});\theta_{\Psi})$};

\path (mnw|-xin)--(xin) node[pos=0.5,above] {$x_i \in \mathcal{D}_s$};
\path (msei|-yout)--(yout) node[pos=0.5,above] {$\mathcal{M}(x_i;\theta)$};

\node[right=0mm of yout.east, anchor=west] (csup) {$\underset{\theta=\{\theta_{\Gamma},\theta_{\Psi}\}}{\min}\enspace \mathcal{C}_{sup}(\mathcal{M}(x_i;\theta), y_i)$};

\draw
node at (0,-45mm) (xin) {}
node [block, right=25mm of xin.east, anchor=west] (l1) {$Layer_1$}
node [block, right=5mm of l1.east, anchor=west] (l2) {$Layer_2$}
node [block, right=5mm of l2.east, anchor=west] (l3) {$Layer_3$}
node [block, right=20mm of l3.east, anchor=west] (l4) {$Layer_4$}
node [right=25mm of l4.east, anchor=west ] (yout) {}
;
\path (l3)--(l4)  coordinate[pos=0.5] (gammaj) node[pos=0.5,below] {$\Gamma(x_j;\theta_{\Gamma})$};

\draw[-](xin) -- node {} (l1);
\draw[-](l1) -- node {} (l2);
\draw[-](l2) -- node {} (l3);
\draw[-](l3) -- node {} (l4);
\draw[->](l4) -- node {} (yout);

\coordinate[above left=3mm of l1.north west] (gammanw);
\coordinate[below right=3mm of l3.south east] (gammase);
\draw [color=gray,thick, dotted] (gammanw) rectangle (gammase);
\node[above=1mm of gammanw, anchor=south west]  (gammalabel) {$\Gamma(\cdot)$};

\coordinate[above left=3mm of l4.north west] (psinw);
\coordinate[below right=3mm of l4.south east] (psise);
\draw [color=gray,thick, dotted] (psinw) rectangle (psise);
\node[above=1mm of psinw, anchor=south west]  {$\Psi(\cdot)$};

\coordinate[above left=3mm of gammalabel.north west] (mnwj);
\coordinate[below right=3mm of psise] (mse);
\draw [color=gray,thick, dashed] (mnwj) rectangle (mse);
\node[above=1mm of mnwj, anchor=south west]  {Replica of $\mathcal{M}(\cdot)$};

\path (mnwj|-xin)--(xin) node[pos=0.5,above] {$\forall x_j \in \mathcal{D}_s$} node[pos=0.5, below] {$j \neq{i}$};
\path (mse|-yout)--(yout) node[pos=0.5,above] {$\mathcal{M}(x_j;\theta)$};

\coordinate  (hinti) at (gammai|-msei);
\coordinate  (hintj) at (gammaj|-mnwj);
\path (hinti)--(hintj) coordinate[pos=0.5] (hint);
\node[] (ch) at (hint) {$\underset{\theta_{\Gamma}}{\min} \enspace \mathcal{C}_h(\Gamma(x_i;\theta_{\Gamma}) ,\Gamma(x_j;\theta_{\Gamma}))$};
\draw[->] (gammai) -- (ch);
\draw[->] (gammaj) -- (ch);

\end{tikzpicture}}
\caption{Constraining the intermediate learned representations to be similar over a decomposed network $\mathcal{M}(\cdot)$ during the \emph{training phase}.}
\label{fig:fig00}
\end{figure*}

Fig.\ref{fig:fig00} illustrates the procedure to measure the dissimilarity in the intermediate representation space $\mathbf{Z}$ between two input samples $x_i$ and $x_j$ with the same label.
Here, we constrained only one hidden layer to be invariant. Extending this procedure for multiple layers is straightforward. It can be done by applying a similar constraint over each concerned layer.

\textbf{Regularized Training Loss}

The full training loss can be formulated as follows:
\begin{align}
  \label{eq:eq12}
  J(\mathcal{D};\theta) &= \underbrace{ \frac{\gamma}{N} \sum_{(x_i,y_i) \in \mathcal{D}} \mathcal{C}_{sup}(\Psi(\Gamma(x_i; \theta_{\Gamma}); \theta_{\Psi}), y_i)}_{\text{Supervised loss } \bm{J_{sup}}} 
  + \underbrace{ \frac{\lambda}{S} \sum_{s=1}^{S} \frac{1}{|\mathcal{D}_s|} \sum_{x_i \in \mathcal{D}_s} J_{h}(x_i; \theta_{\Gamma})}_{\text{Hint penalty }\bm{J_{H}}}
\end{align}

\noindent where $\gamma$  and $\lambda$ are regularization weights, $\mathcal{C}_{sup}(\cdot, \cdot)$ the classification loss function.
If one use a  dissimilarity measure $\mathcal{C}_h(\cdot, \cdot)$ in $J_h$ that is symmetrical such as typically a distance, summations in the term $\bm{J_H}$ could be rewritten to prevent the same sample couple to appear twice.

Eq.\ref{eq:eq12} shares a similarity with the contrastive loss \cite{chopraHL05, hadsellCL06, bromley1993}. This last one is composed of two terms. One term constrains the learned model to project similar inputs to be closer in the output space. In Eq.\ref{eq:eq12}, this is represented by the hint term. In \cite{chopraHL05, hadsellCL06, bromley1993}, to avoid collapsing all the inputs into one single output point, the contrastive loss uses a second term which projects dissimilar points far from each other by at least a minimal distance. In Eq.\ref{eq:eq12}, the supervised term prevents, implicitly, this collapsing by constraining the extracted representations to be discriminative with respect to each class in order to minimize the classification training error.

\subsection{Implementation and Optimization Details}
\label{implandoptim}

In the present work, we have chosen the cross-entropy as the classification loss $\mathcal{C}_{sup}(\cdot, \cdot)$.

In order to quantify how much two representation vectors in $\mathbf{Z}$ are dissimilar we proceed using a distance based approach for $\mathcal{C}_h(\cdot, \cdot)$.
We study three different measures: the squared Euclidean distance (\emph{SED}),

\begin{equation}
	\label{eq:eq2}
	\mathcal{C}_h(a, b) = \lVert{a} - {b}\rVert^2_2 =  \sum_{v=1}^V ({a}_{ v} - {b}_{v})^2\enspace,
\end{equation}

\noindent the normalized Manhattan distance (\emph{NMD}),

\begin{equation}
	\label{eq:eq22}
	\mathcal{C}_h({a}, {b}) = \frac{1}{V}\sum_{v=1}^V |{a}_{v} - {b}_{v}|\enspace,
\end{equation}

\noindent and the angular similarity (\emph{AS}),

\begin{equation}
\label{eq:eq23}
\mathcal{C}_h({a}, {b}) = \arccos \left(\frac{\langle\, {a}, {b}\rangle}{\lVert {a}\rVert_2 \; \lVert {b}\rVert_2}\right) \enspace.
\end{equation}

Minimizing the loss function of Eq.\ref{eq:eq12} is achieved using Stochastic Gradient Descent (SGD). Eq.\ref{eq:eq12} can be seen as multi-tasking where two tasks represented by the supervised term and the hint term are in concurrence. One way to minimize Eq.\ref{eq:eq12} is to perform a parallel optimization of both tasks by adding their gradient. Summing up the gradient of both tasks can lead to issues mainly because both tasks have different objectives that do not steer necessarily in the same direction. In order to avoid these issues, we propose to separate the gradients by alternating between the two terms at each mini-batch which showed to work well in practice \cite{caruana97ML, weston2012, collobert08ICML, bel16}. Moreover, we use two separate optimizers where each term has its own optimizer. By doing so, we make sure that both gradients are separated.

On a large dataset, computing all the dissimilarity measures in $\bm{J_H}$ in Eq.\ref{eq:eq12} over the whole training dataset is computationally expensive due to the large number of pairs. Therefore, we propose to compute it only over the mini-batch presented to the network. Consequently, we need to shuffle the training set $\mathcal{D}$ periodically in order to ensure that the network has seen almost all the possible combinations of the pairs. We describe our implementation in Alg.\ref{alg:alg0}.

\begin{algorithm}[h!]
    \caption{Our training strategy}
    \label{alg:alg0}
    \begin{algorithmic}[1]
      \State $\mathcal{D}$ is the 
      training set. $B_s$ a mini-batch. $B_r$ a mini-batch of all the possible pairs in $B_s$ (Eq.\ref{eq:eq12}). $OP_s$ an optimizer of the supervised term. $OP_r$ an optimizer of the dissimilarity term. \texttt{max\_epochs}: maximum epochs. $\gamma, \lambda$ are regularization weights.
      \For{i=1..\texttt{max\_epoch}}
      \State Shuffle $\mathcal{D}$. Then, split it into mini-batches.
      \For{$(B_s, B_r)$ in $\mathcal{D}$}
        \State Make a gradient step toward $\bm{J_{sup}}$ using $B_s$ and \par \hskip\algorithmicindent $OP_s$. (Eq.\ref{eq:eq12})
        \State Make a gradient step toward $\bm{J_H}$ using $B_h$ and \par \hskip\algorithmicindent $OP_r$. (Eq.\ref{eq:eq12})
      \EndFor
      \EndFor
    \end{algorithmic}
\end{algorithm}

\section{Experiments}
\label{exps}

In this section, we evaluate our regularization framework for training deep networks on a classification task as described in Section \ref{method}. In order to show the effect of using our regularization on the generalization performance, we will mainly compare the generalization error of a network trained with and without our regularizer on different benchmarks of classification problems.

\subsection{Classification Problems and Experimental Methodology}

\begin{figure}[!ht]
\centering
\includegraphics[scale=0.3]{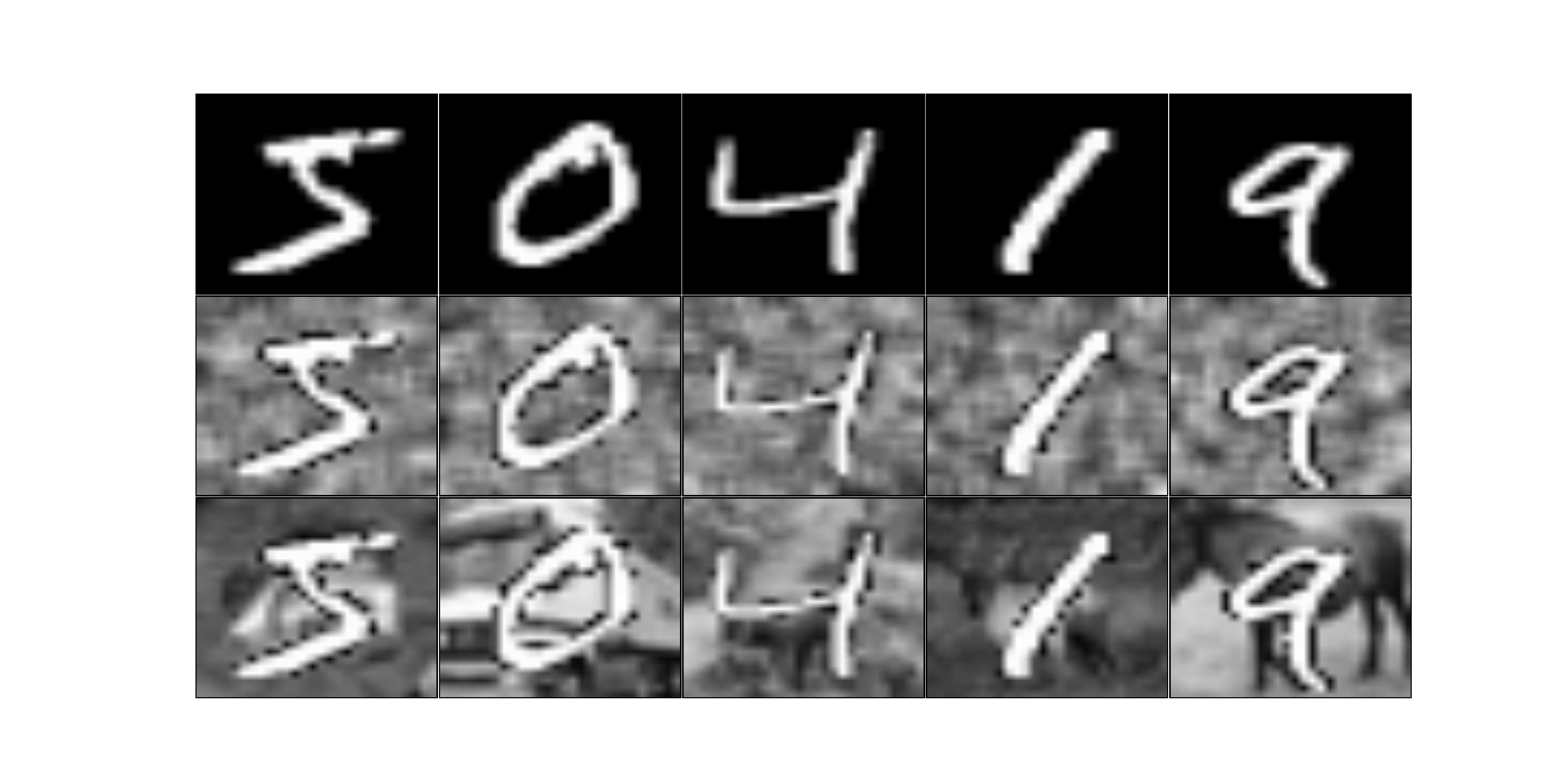}
\caption{Samples from training set of each benchmark. \emph{Top row}: \emph{mnist-std} benchmark. \emph{Middle row}: \emph{mnist-noise} benchmark. \emph{Bottom row}: \emph{mnist-img} benchmark.}
\label{fig:figsamples}
\end{figure}

In our experiments, we consider three classification problems. We start by the standard MNIST digit dataset. Then, we complicate the classification task by adding different types of noise. We consider the three following problems:

\begin{itemize}
\item The standard MNIST digit classification problem with $\mathit{50000}$, $\mathit{10000}$ and $\mathit{10000}$ training, validation and test set. We refer to this benchmark as \emph{mnist-std}. (Fig.\ref{fig:figsamples}, top row).
\item MNIST digit classification problem where we use a background mask composed of a random noise followed by a uniform filter. The dataset is composed of $\mathit{100000}$, $\mathit{20000}$ and $\mathit{50000}$ samples for train, validation and test set. Each set is generated from the corresponding set in the benchmark \emph{mnist-std}. We refer to this benchmark as \emph{mnist-noise}. (Fig.\ref{fig:figsamples}, middle row).
\item MNIST digit classification problem where we use a background mask composed of a random picture taken from CIFAR-10 dataset \cite{krizhevsky09learningmultiple}. This benchmark is composed of $\mathit{100000}$ samples for training built upon $\mathit{40000}$ training samples of CIFAR-10 training set, $\mathit{20000}$ samples for validation built upon the rest of CIFAR-10 training set (i.e. $\mathit{10000}$ samples) and $\mathit{50000}$ samples for test built upon the $\mathit{10000}$ test samples of CIFAR-10. We refer to this benchmark as \emph{mnist-img}. (Fig.\ref{fig:figsamples}, bottom row).
\end{itemize}

All the images are $28 \times 28$ gray-scale values scaled to $[0, 1]$. In order to study the behavior of our proposal where we have few training samples, we use different configurations for the training set size. We consider four configurations where we take only $\mathit{1000}$, $\mathit{3000}$, $\mathit{5000}$, $\mathit{50000}$ or $\mathit{100000}$ training samples from the whole available training set. We refer to each configuration by $\mathit{1k}$, $\mathit{3k}$, $\mathit{5k}$, $\mathit{50k}$ and $\mathit{100k}$ respectively. For the benchmark \emph{mnist-std}, only the configurations $\mathit{1k}$, $\mathit{3k}$, $\mathit{5k}$ and $\mathit{50k}$ are considered.

For all the experiments, we consider the two following neural network architectures:
\begin{itemize}
\item Multilayer perceptron with 3 hidden layers followed by a classification output layer. We use the same architecture as in \cite{devriesESANN2016} which is $1200-1200-200$. This model is referred to as \emph{mlp}.
\item LeNet convolutional network \cite{lecun98gradient}, which is well known in computer vision tasks, (with similar architecture to LeNet-4) with 2 convolution layers with 20 and 50 filters of size $5 \times 5$, followed by a dense layer of size $500$, followed by a classification output layer. This model is referred to as \emph{lenet}.
\end{itemize}
Each model has three hidden layers, we refer to each layer from the input toward the output layer by: $h_1, h_2$ and $h_3$ respectively. The output layer is referred to as $h_4$. When using our hint term, we refer to the model by \emph{mlp + hint} and \emph{lenet + hint} for the \emph{mlp} and \emph{lenet} models respectively.

Each experiment is repeated $7$ times. The best and the worst test classification error cases are discarded. We report the \emph{mean} $\pm$ \emph{standard deviation} of the validation (vl) and the test (tst) classification error of each benchmark. Models without regularization are trained for $\mathit{2000}$ epochs. All the models regularized with our proposal are trained for $\mathit{400}$ epochs which we found enough to converge and find a better model over the validation set. All the trainings are performed using stochastic gradient descent with an adaptive learning rate applied using AdaDelta \cite{zeiler12}, with a batch size of $\mathit{100}$.

\textbf{Technical Details}:

\begin{itemize}
\item We found that layers with bounded activation functions such as the logistic sigmoid or the hyperbolic tangent function are more suitable when applying our hint term. Applying the regularization term over a layer with unbounded activation function such as the Relu \cite{NairHicml10} did not show an improvement.
\item In practice, we found that setting $\gamma=1, \lambda=1$ works well.
\end{itemize}
The source code of our implementation is freely available \footnote{\url{https://github.com/sbelharbi/learning-class-invariant-features}}.

\subsection{Results}
As we have described in Sec.\ref{method}, our hint term can be applied at any hidden layer of the network. In this section, we perform a set of experiments in order to have an idea which one is more adequate to use our regularization. To do so, we trained the \emph{mlp} model for classification task over the benchmark \emph{mnist-std} using different configurations with and without regularization. The regularization is applied for one hidden layer at a time $h_1, h_2$ or $h_3$. We used the squared Euclidean distance (Eq.\ref{eq:eq2}) as a dissimilarity measure. The obtained results are presented in Tab.\ref{tab:tab17}.

\begin{table*}[!htbp]
    \centering
  \resizebox{1.\textwidth}{!}{
   \begin{tabular}{|l||c|c||c|c||c|c||c|c|}
		\hline
        Model/train data size&\multicolumn{2}{|c||}{$\mathit{1k}$}&\multicolumn{2}{c||}{$\mathit{3k}$}&\multicolumn{2}{c||}{$\mathit{5k}$}&\multicolumn{2}{c|}{$\mathit{50k}$}\\        
                \hline

                \multicolumn{1}{c||}{}
                &vl&tst&
                vl&tst&
                vl&tst&
                vl&tst\\
                \cline{2-9}
                \cline{2-9}
                \multicolumn{1}{c||}{}&\multicolumn{8}{|c|}{\textbf{\emph{mlp}}}\\
                \cline{2-9}
                \cline{2-9}
                \multicolumn{1}{c||}{}&
                $10.49\pm0.031$&
                $11.24\pm0.050$&
                $6.69\pm0.039$&
                $7.17\pm0.010$&
                $5.262\pm0.030$&
                $5.63\pm0.126$&
                $\bm{1.574\pm0.016}$&
                $\bm{1.66\pm0.016}$\\
                \cline{2-9}
                \cline{2-9}
                \multicolumn{1}{c||}{}&\multicolumn{8}{|c|}{\textbf{\emph{mlp + reg.}}}\\
                \cline{2-9}
                \hline
                $h_3$&
                $\bm{8.80\pm0.093}$&
                $\bm{9.50\pm0.093}$&
                $\bm{5.81\pm0.104}$&
                $\bm{6.24\pm0.069}$&
                $\bm{4.74\pm0.065}$&
                $\bm{5.05\pm0.035}$&
                $1.67\pm0.043$&
                $1.73\pm0.080$\\
                \hline
                \hline
                $h_2$&
                $11.48\pm0.081$&
                $12.32\pm0.090$&
                $6.72\pm0.031$&
                $7.29\pm0.038$&
                $5.33\pm0.031$&
                $5.84\pm0.030$&
                $1.88\pm0.043$&
                $1.97\pm0.071$\\
                \hline
                \hline
                $h_1$&
                $12.15\pm0.043$&
                $12.74\pm0.189$&
                $6.75\pm0.041$&
                $7.26\pm0.049$&
                $5.35\pm0.028$&
                $5.87\pm0.050$&
                $1.83\pm0.033$&
                $1.95\pm0.025$\\
                \hline
	\end{tabular}
        }
	\caption{\footnotesize Mean $\pm$ standard deviation error over validation and test set of the benchmark \emph{mnist-std} using the model \emph{mlp} and the SED as dissimilarity measure over the different hidden layers: $h_1, h_2, h_3$. (\textbf{bold font indicates lowest error.})}
	\label{tab:tab17}
\end{table*}

From Tab.\ref{tab:tab17}, one can see that regularizing low layers $h_1, h_2$ did not help improving the performance error but it did increase it in the configuration $1k$, for instance. This may be explained by the fact that low layers in neural networks tend to learn low representations which are \emph{shared} among high representations. This means that these representations are not ready yet to discriminate between the classes. Therefore, they can not be used to describe each class separately. This makes our regularization inadequate at these levels because we aim at constraining the representations to be similar within each class while these layers are incapable to deliver such representations. Therefore, regularizing these layers may hamper their learning. As a future work, we think that it would be beneficial to use at low layers a regularization term that constrains the representations of samples within different classes be dissimilar such as the one in the contrastive loss \cite{chopraHL05, hadsellCL06, bromley1993}.

In the case of regularizing the last hidden layer $h_3$, we notice from Tab.\ref{tab:tab17} an important improvement in the classification error over the validation and the test set in most configurations. This may be explained by the fact that the representations at this layer are more abstract, therefore, they are able to discriminate the classes. Our regularization term constrains these representations to be tighter by re-enforcing their invariance which helps in generalization. Therefore, applying our hint term over the last hidden layer makes more sense and supports the idea that high layers in neural networks learn more abstract representations. Making these discriminative representations invariant helps the linear output layer in the classification task. For all the following experiments, we apply hint term over the last hidden layer.
Moreover, one can notice that our regularization has less impact when adding more training samples. For instance, we reduced the classification test error by: $1.74\%$, $0.92\%$ and $0.58\%$ in the configurations $1k$, $3k$ and $5k$. This suggests that our proposal is more efficient in the case where few training samples are available. However, this does not exclude using it for large training datasets as we will see later (Tab.\ref{tab:tab18}, \ref{tab:tab20}). We believe that this behavior depends mostly on the model's capacity to learn invariant representations. For instance, from the invariance perspective, convolutional networks are more adapted, conceptually, to process visual content than multilayers perceptrons.

\bigskip

In another experimental setup, we investigated the effect of the measure used to compute the dissimilarity between two feature vectors as described in Section.\ref{implandoptim}. To do so, we applied our hint term over the last hidden layer $h_3$ using the measures \emph{SED}, \emph{NMD} and \emph{AS} over the benchmark \emph{mnist-std}. The obtained results are presented in Tab.\ref{tab:tab18}. These results show that the squared Euclidean distance performs significantly better than the other measures and has more stability when changing the number of training samples ($\mathit{1k}$, $\mathit{3k}$, $\mathit{5k}$, $\mathit{50k}$) or the model (\emph{mlp}, \emph{lenet}).

\begin{table*}[!htbp]
    \centering
  \resizebox{1.\textwidth}{!}{
   \begin{tabular}{|l||c|c||c|c||c|c||c|c|}
		\hline
        Model/train data size&\multicolumn{2}{|c||}{$\mathit{1k}$}&\multicolumn{2}{c||}{$\mathit{3k}$}&\multicolumn{2}{c||}{$\mathit{5k}$}&\multicolumn{2}{c|}{$\mathit{50K}$}\\        
                \hline
                \multicolumn{1}{c||}{}
                &vl&tst&
                vl&tst&
                vl&tst&
                vl&tst\\
                \cline{2-9}
                \multicolumn{1}{c||}{}
                &\multicolumn{8}{|c|}{\textbf{MLP}}\\
                \hline{2-9}
                \emph{mlp}&
                $10.49\pm0.031$&
                $11.24\pm0.050$&
                $6.69\pm0.039$&
                $7.17\pm0.010$&
                $5.262\pm0.030$&
                $5.63\pm0.126$&
                $1.574\pm0.016$&
                $1.66\pm0.016$\\
                \hline
                \hline
                \emph{mlp + hint} (SED)&
                $\bm{8.80\pm0.093}$&
                $\bm{9.50\pm0.093}$&
                $\bm{5.81\pm0.104}$&
                $\bm{6.24\pm0.069}$&
                $\bm{4.74\pm0.065}$&
                $\bm{5.05\pm0.035}$&
                $1.67\pm0.043$&
                $1.73\pm0.080$\\
                \hline
                \hline
                \emph{mlp + hint} (NMD)&
                $10.32\pm0.028$&
                $10.92\pm0.094$&
                $6.69\pm0.075$&
                $7.22\pm0.059$&
                $5.34\pm0.035$&
                $5.79\pm0.045$&
                $1.44\pm0.020$&
                $1.47\pm0.020$\\
                \hline
                \hline
                \emph{mlp + hint} (AS)&
                $10.27\pm0.068$&
                $10.71\pm0.123$&
                $6.52\pm0.044$&
                $6.89\pm0.013$&
                $4.96\pm0.041$&
                $5.25\pm0.051$&
                $\bm{1.37\pm0.023}$&
                $\bm{1.37\pm0.025}$\\
                \hline
                \multicolumn{1}{c||}{}
                &\multicolumn{8}{|c|}{\textbf{Lenet}}\\
                \hline
                \hline
                \emph{lenet}&
                $6.25\pm0.016$&
                $7.27\pm0.033$&
                $3.65\pm0.085$&
                $4.02\pm0.073$&
                $2.62\pm0.031$&
                $2.90\pm0.058$&
                $1.31\pm0.028$&
                $1.23\pm0.024$\\
                \hline
                \hline
                \emph{lenet + hint} (SED)&
                $\bm{4.54\pm0.150}$&
                $5.05\pm0.115$&
                $\bm{2.70\pm0.124}$&
                $\bm{2.85\pm0.082}$&
                $\bm{2.06\pm0.113}$&
                $\bm{2.37\pm0.105}$&
                $\bm{0.97\pm0.087}$&
                $\bm{1.04\pm0.060}$\\
                \hline
                \hline
                \emph{lenet + hint} (NMD)&
                $6.70\pm0.040$&
                $\bm{4.60\pm0.065}$&
                $3.85\pm0.032$&
                $4.30\pm0.036$&
                $2.87\pm0.045$&
                $3.14\pm0.035$&
                $1.99\pm0.043$&
                $2.075\pm0.079$\\
                \hline
                \hline
                \emph{lenet + hint} (AS)&
                $6.72\pm0.024$&
                $7.66\pm0.024$&
                $3.86\pm0.049$&
                $4.26\pm0.049$&
                $2.80\pm0.033$&
                $3.12\pm0.021$&
                $1.75\pm0.123$&
                $1.97\pm0.063$\\
                \hline
	\end{tabular}
        }
	\caption{\footnotesize Mean $\pm$ standard deviation error over validation and test set of the benchmark \emph{mnist-std} using different dissimilarity measures (\emph{SED}, \emph{NMD}, \emph{AS}) over the layer $h_3$. (\textbf{bold font indicates lowest error.})}
	\label{tab:tab18}
\end{table*}

\bigskip

In another experiment, we evaluated the benchmarks \emph{mnist-noise} and \emph{mnist-img}, which are more difficult compared to \emph{mnist-std}, using the model \emph{lenet} which is more suitable to process visual content. Similarly to the previous experiments, we applied our regularization term over the last hidden layer $h_3$ using the SED measure. The results depicted in Tab.\ref{tab:tab20} show again that using our proposal improves the generalization error of the network particularly when only few training samples are available. For example, our regularization allows to reduce the classification error over the test set by $2.98\%$ and by $4.16\%$ over the benchmark \emph{mnist-noise} and \emph{mnist-img}, respectively when using only $\mathit{1k}$ training samples.

\begin{table*}[!htbp]
    \centering
  \resizebox{1.\textwidth}{!}{
   \begin{tabular}{|l||c|c||c|c||c|c||c|c|}
		\hline
        Model/train data size&\multicolumn{2}{|c||}{$\mathit{1k}$}&\multicolumn{2}{c||}{$\mathit{3k}$}&\multicolumn{2}{c||}{$\mathit{5k}$}&\multicolumn{2}{c|}{$\mathit{100k}$}\\        
                \hline
                \multicolumn{1}{c||}{}
                &vl&tst&
                vl&tst&
                vl&tst&
                vl&tst\\
                \cline{2-9}
                \multicolumn{1}{c||}{}
                &\multicolumn{8}{|c|}{\textbf{\emph{mnist-noise}}}\\
                \hline
                \emph{lenet}&
                $9.62\pm0.123$&
                $10.72\pm0.116$&
                $5.95\pm0.059$&
                $6.39\pm0.032$&
                $4.92\pm0.036$&
                $5.11\pm0.012$&
                $1.90\pm0.020$&
                $2.011\pm0.018$\\
                \hline
                \hline
                \emph{lenet + hint}&
                $\bm{7.12\pm0.200}$&
                $\bm{7.74\pm0.148}$&
                $\bm{4.09\pm0.130}$&
                $\bm{4.62\pm0.059}$&
                $\bm{3.53\pm0.117}$&
                $\bm{3.98\pm0.167}$&
                $\bm{1.60\pm0.107}$&
                $\bm{1.64\pm0.116}$\\
                \hline
                \multicolumn{1}{c||}{}
                &\multicolumn{8}{|c|}{\textbf{\emph{mnist-img}}}\\
                \hline
                \emph{lenet}&
                $13.88\pm0.114$&
                $15.34\pm0.124$&
                $8.34\pm0.030$&
                $8.66\pm0.024$&
                $6.64\pm0.057$&
                $6.46\pm0.033$&
                $2.53\pm0.080$&
                $2.55\pm0.007$\\
                \hline
                \hline
                \emph{lenet + hint}&
                $\bm{10.30\pm0.425}$&
                $\bm{11.18\pm0.290}$&
                $\bm{6.19\pm0.281}$&
                $\bm{6.61\pm0.212}$&
                $\bm{5.37\pm0.358}$&
                $\bm{5.65\pm0.310}$&
                $\bm{2.15\pm0.105}$&
                $\bm{2.21\pm0.032}$\\
                \hline
	\end{tabular}
        }
	\caption{\footnotesize Mean $\pm$ standard deviation error over validation and test set of the benchmarks \emph{mnist-noise} and \emph{mnist-img} using \emph{lenet} model (regularization applied over the layer $h_3$). (\textbf{bold font indicates lowest error.})}
	\label{tab:tab20}
\end{table*}

\bigskip

Based on the above results, we conclude that using our hint term in the context of classification task using neural networks is helpful in improving their generalization error particularly when only few training samples are available. This generalization improvement came at the price of an extra computational cost due the dissimilarity measures between pair of samples. Our experiments showed that regularizing the last hidden layer using the squared Euclidean distance give better results. More generally, the obtained results confirm that guiding the learning process of the intermediate representations of a neural network can be helpful to improve its generalization.

\subsection{On Learning Invariance within Neural Networks}
\begin{figure*}[!htbp]
\centering
\includegraphics[scale=0.3]{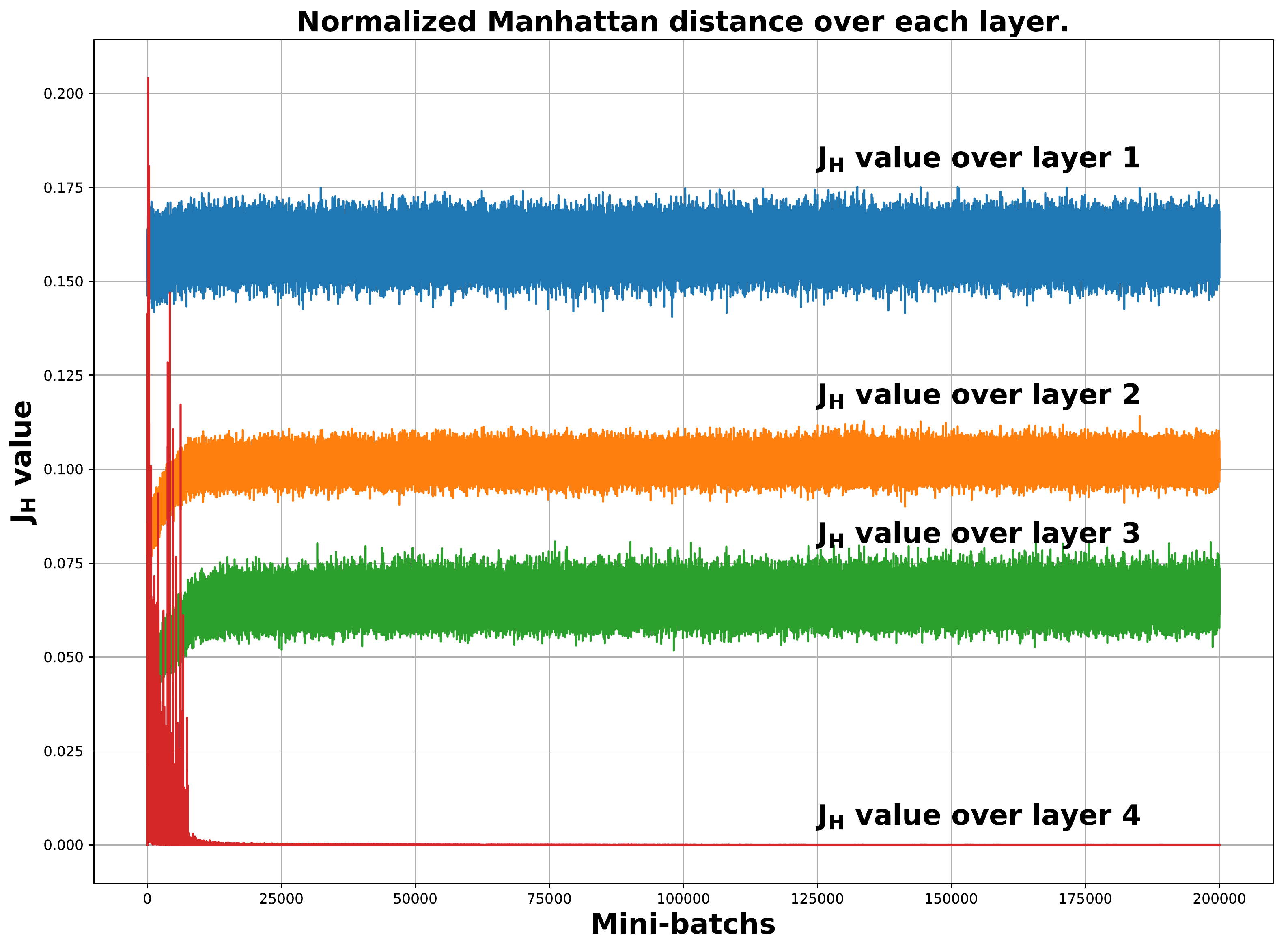}
\caption{Measuring the hint term $\bm{J_H}$ of Eq.\ref{eq:eq12} over the training set within each layer (simultaneously) of the \emph{mlp} over the train set of \emph{mnist-std} benchmark for a binary classification task: the digit \quotes{1} against the digit \quotes{7}.}
\label{fig:figmeasuremlpandlenet}
\end{figure*}

We show in this section an intriguing property of the learned representations at each layer of a neural network from the invariance perspective. For this purpose and for the sake of simplicity, we consider a binary classification case of the two digits \quotes{1} and \quotes{7}. Furthermore, we consider the \emph{mlp} model over the \emph{lenet} in order to be able to measure the features invariances over all the layers. We trained the \emph{mlp} model over the benchmark \emph{mnist-std} where we used all the available training samples of both digits. The model is trained without our regularization. However, we tracked, at each layer and at the same time, the value of the hint term $\bm{J_H}$ in Eq.\ref{eq:eq12} over the training set using the normalized Manhattan distance as a dissimilarity measure. This particular dissimilarity measure allows comparing the representations invariance between the different layers due to the normalization of the measure by the representations dimension. The obtained results are depicted in Fig.\ref{fig:figmeasuremlpandlenet} where the x-axis represents the number of mini-batches already processed and the y-axis represents the value of the hint term $\bm{J_H}$ at each layer. Low value of $\bm{J_H}$ means high invariance (better case) whereas high value of $\bm{J_H}$ means low invariance.

In Fig.\ref{fig:figmeasuremlpandlenet}, we note two main observations:
\begin{itemize}
\item The value of the hint term $\bm{J_H}$ is reduced through the depth of the network which means that the network learns more invariant representations at each layer in this order: layer 1, 2, 3, 4. This result supports the idea that abstract representations, which are known to be more invariant, are learned toward the top layers.
\item At each layer, the network does not seem to learn to improve the invariance of the learned representations by reducing $\bm{J_H}$. It appears that the representations invariance is kept steady all along the training process. Only the output layer has learned to reduce the value of $\bm{J_H}$ term because minimizing the classification term $\bm{J_{sup}}$ reduces automatically our hint term $\bm{J_H}$. This shows a flaw in the back-propagation procedure with respect to learning intermediate representations. Assisting the propagated error through regularization can be helpful to guide the hidden layers to learn more suitable representations.
\end{itemize}

These results show that relying on the classification error propagated from the output layer does not necessarily constrain the hidden layers to learn better representations for classification task. Therefore, one would like to use different prior knowledge to guide the internal layers to learn better representations which is our future work. Using these guidelines can help improving neural networks generalization especially when trained with few samples.

\section{Conclusion}
\label{conclusion}
We have presented in this work a new regularization framework for training neural networks for classification task. Our regularization constrains the hidden layers of the network to learn class-wise invariant representations where samples of the same class have the same  representation. Empirical results over MNIST dataset and its variants showed that the proposed regularization helps neural networks to generalize better particularly when few training samples are available which is the case in many real world applications.

Another result based on tracking the representation invariance within the network layers confirms that neural networks tend to learn invariant representations throughout staking multiple layers. However, an intriguing observation is that the invariance level does not seem to be improved, within the same layer, through learning. We found that the hidden layers tend to maintain a certain level of invariance through the training process.

All the results found in this work suggest that guiding the learning process of the internal representations of a neural network can be helpful to train them and improve their generalization particularly when few training samples are available. Furthermore, this shows that the classification error propagated from the output layer does not necessarily train the hidden layers to provide better representations. This encourages us to explore other directions to incorporate different prior knowledge to constrain the hidden layers to learn better representations in order to improve the generalization of the network and be able to train it with less data.

\subsubsection*{Acknowledgments}

This work has been partly supported by the grant ANR-16-CE23-0006 \quotes{Deep in France} and benefited from computational means from 
CRIANN, the contributions of which are greatly appreciated.


\bibliography{bibliography}  





\end{document}